\documentclass[twoside,11pt]{article}

\usepackage{jmlr2e}
\usepackage[utf8]{inputenc} % allow utf-8 input
\usepackage[T1]{fontenc}    % use 8-bit T1 fonts
\usepackage{hyperref}       % hyperlinks
\usepackage{url}            % simple URL typesetting
\usepackage{booktabs}       % professional-quality tables
\usepackage{amsfonts}       % blackboard math symbols
\usepackage{mathtools}       % blackboard math symbols
\usepackage{nicefrac}       % compact symbols for 1/2, etc.
\usepackage{microtype}      % microtypography

\newcommand{\indep}{\perp \!\!\! \perp}

\ShortHeadings{Higher-Order Generalization Bounds}{Warrell and Gerstein}
\firstpageno{1}

\begin{document}

\title{Higher-Order Generalization Bounds: Learning Deep Probabilistic Programs via PAC-Bayes Objectives}

\author{\name Jonathan Warrell \email jonathan.warrell@yale.edu \\
       \addr Program in Computational Biology and Bioinformatics \\
	   Department of Molecular Biophysics and Biochemistry \\
       Yale University, New Haven, CT, 06520, USA
       \AND
       \name Mark Gerstein \email pi@gersteinlab.org \\
       \addr Program in Computational Biology and Bioinformatics \\
	   Department of Molecular Biophysics and Biochemistry \\
       Yale University, New Haven, CT, 06520, USA}

\editor{}

\maketitle

\begin{abstract}%   <- trailing '%' for backward compatibility of .sty file
Deep Probabilistic Programming (DPP) allows powerful models based on recursive computation to be learned using efficient deep-learning optimization techniques. Additionally, DPP offers a unified perspective, where inference and learning algorithms are treated on a par with models as stochastic programs. Here, we offer a framework for representing and learning flexible PAC-Bayes bounds as stochastic programs using DPP-based methods. In particular, we show that DPP techniques may be leveraged to derive generalization bounds that draw on the compositionality of DPP representations. In turn, the bounds we introduce offer principled training objectives for higher-order probabilistic programs.  We offer a definition of a higher-order generalization bound, which naturally encompasses single- and multi-task generalization perspectives (including transfer- and meta-learning) and a novel class of bound based on a learned measure of model complexity.  Further, we show how modified forms of all higher-order bounds can be efficiently optimized as objectives for DPP training, using variational techniques.  We test our framework using single- and multi-task generalization settings on synthetic and biological data, showing improved performance and generalization prediction using flexible DPP model representations and learned complexity measures.
\end{abstract}


\section{Introduction}

Deep Probabilistic Programming (DPP) provides a framework which combines the benefits of models based on recursion with the strengths of deep-learning based optimization [21,22].  In addition, it provides a general model of probabilistic computation, for which recent work has provided associated operational and denotational semantics [11,18,20].  As such, there is no need to make a firm distinction between models and inference/training algorithms in a DPP setting, since all are stochastic functions; this perspective naturally leads to treating inference/training optimization algorithms as `first-class' citizens which may themselves be optimized [21,22].  However, the objectives used to learn DPPs are typically standard functions such as the log-likelihood, or evidence lower bound (ELBO) [21,22].  Potentially, more sophisticated training objectives can be derived by considering generalization bounds linked to specific learning settings, for instance, transfer- or meta-learning [4,15,25], or data-dependent priors [3,7,8,14,17].   We propose that, as for optimization algorithms, generalization bounds should themselves be viewed as `first-class' stochastic functions to be learned in a DPP setting.  Doing so has the potential both to motivate new bounds based on DPP compositionality and data-dependence, and to motivate new objectives for training programs and collections of programs in a DPP setting.

In light of the above, we propose a general definition of a {\em higher-order (h-o) generalization bound}.  Briefly, this is a higher-order function that returns a generalization bound, which holds with high-probability given independence conditions regarding its inputs, and those of the bound returned.  We show that this definition provides a unified representation for existing and novel bounds for single- and multi-task generalization based on a PAC-Bayes framework.  Particularly, we show that existing transfer-/meta-learning [4,15,25] and data-dependent [3,7,8,14,17] PAC-Bayes bounds can be formulated directly as h-o bounds.  An advantage of the DPP perspective is that it places minimal restrictions on the functional forms used by these bounds;  existing approaches focus on restricted forms of the prior and/or posterior [3,4,7,8,14,15,17,25,26], or hyper-prior and -posterior in the meta-learning setting [4,15,25], which in our framework may be arbitrary in form.  We show that modified forms of these bounds can be derived in our framework using recent variational and multi-sample techniques [16,19,21,22], providing objectives that can be efficiently optimized in the DPP setting.

Further, we introduce a novel form of h-o bound which uses a learned measure of model complexity (which we term a second-order complexity bound).  This bound takes as input a {\em generalization classifier}, which predicts a model's task-specific generalization error from its parameters, and may be trained jointly with a base-level task classifier.  The bound may be naturally extended to a meta-learning setting using the framework we introduce.  Unlike previous meta-learning PAC-Bayes bounds [4,15,25], our learned model complexity bound uses a feed-forward model to estimate the per-task model complexity, and hence avoids the need to estimate separate KL-divergence terms for each task during training.  As above, we show that variational and multi-sample techniques can be used to derive a tractable training objective from our novel bound.  Finally, we introduce a convenient stochastic type system for expressing DPP models and h-o bounds, which directly adapts aspects of the systems introduced in [11,14,18,20] to provide a framework with a clear underlying semantics.

We begin by outlining our stochastic type-system, before introducing our general framework for h-o generalization bounds.  We then give several examples of how existing transfer- and meta-learning bounds may be modified to produce DPP variational objectives in our framework, before introducing our second-order complexity bound.  We test our framework using single- and multi-task generalization settings on synthetic and biological data, the latter consisting of gene expression data from a psychiatric genomics dataset, containing subjects with and without related psychiatric disorders [9,23].  We show improved task performance and generalization prediction using the flexible DPP objectives and learned complexity measure we introduce.

\section{Preliminaries: Stochastic Type System}

We first introduce a notation for stochastic types and programs, which we will use throughout the paper.  The system we introduce is a variant of the higher-order language specified in [20], with some small differences that we note.  We assume we have types $A,B,C...,Z$ along with function and product types (e.g. $A\rightarrow B$ and $A\times B$ resp.), and write $a:A$ for $a$ belongs to type $A$.  The type $I$ denotes the unit interval, and we write $A'$ for the type of distributions over $A$, where we assume for convenience all types are discrete, and $A'$ is the subset of $(A\rightarrow I)$ containing only maps which sum to 1.  $A'$ is thus equivalent to $(\text{Mass}\; A)$ in [18] (the type $A$ is mapped to by the {\em Mass Function Monad}).  Further, we include the constructions $\text{sample}()$ and $\text{thunk}()$ as used in the language of [20].  For $p:A'$, $\text{sample}(p)$ is a term of type $A$, which reduces to a base term in $A$ through probabilistic $\beta$-reduction with probability $p(a)$ (we assume a probabilistic reduction semantics of the kind in [20]).   Further, for $a:A$, we let $\text{thunk}(a)$ denote a term of type $A'$; specifically, $\text{thunk}(a)$ is a function which, for a term $a'$, returns the probability that $a$ reduces to $a'$.  Hence $(\text{thunk}(a))(a')=P(a\rightarrow_{\beta} a')$, where $\rightarrow_{\beta}$ denotes probabilistic reduction.  Particularly, we have that $\text{thunk}(\text{sample}(p))=p$, and $\text{sample}(\text{thunk}(a))=a$.  Here, `thunking' can be viewed as a means of suspending a probabilistic program $\text{sample}(p)$ so that it is prevented from executing,  which may be reactivated by applying a sampling statement.  We note that, unlike [20], we do not distinguish between distributions and thunked programs over a given type $A$, since we assign them both to $A'$, and hence `sampling' and `forcing' are synonyms in our system.  We outline further technical details of our system and its relation to [20] in Appendix \ref{app:A}.

For convenience, we now introduce a concise notation that will allow us to express DPPs and associated bounds efficiently throughout the paper.  We first introduce the notation informally.  For a suspended probabilistic program, $p:A'$, we introduce a hierarchy of sampling statements, where $p^*=\text{sample}(p)$, and $p^+$, $p^{++}$, $p^{+++}$..., represent programs suspended at varying levels of execution (referred to as sampling levels 1, 2, 3...).  Specifically, if $f:A\times A \rightarrow B$, then we may write $f(p^+,a'')$ for a suspended program of type $B'$, which when run (i.e. sampled), first converts $p^+$ to $p^*$, which reduces probabilistically to  $a'$, and then applies $f$ to $(a',a'')$ to generate $b$.  Further, we may write $f(p^+,p^{++})$ for a higher-order suspended program of type $B''$; here, the rule is that running a statement (sampling and reducing) converts all sampling statements at level 1 (i.e. of the form $p^+$) to $p^*$ statements, and decrements the sampling levels of all others by one, before applying probabilistic reduction.  Hence, $\text{sample}(f(p^+,p^{++})) \rightarrow_{\beta} f(p^*,p^+)$, and $f(p^+,p^{++}):B''$.  Formally, this notation is shorthand for the following construction:
\begin{eqnarray}\label{eq:typeSystem1}
f[p_1^+,p_2^+,...] &=& \text{thunk}(\lambda (a_1,a_2,...).f^{(-)}[a/p_1^+,a/p_2^+,...]\; (p_1^*,p_2^*,...))
\end{eqnarray}
where $f[a_1,a_2,...]$ denotes an expression $f$ containing $a_1,a_2,...$, $f[a_1/b_1,a_2/b_2,...]$ denotes the result of replacing $b_1$ with $a_1$, $b_2$ with $a_2$ etc. in $f$, and $f^{(-)}$ denotes the result of decrementing the sampling levels of all statements by one in $f$ (following any substitutions; note that our notation can be further developed to incorporate {\em memoization} as in [10], see Appendix \ref{app:A}).  The rule in Eq. \ref{eq:typeSystem1} can be recursively applied to the above example: $f(p^+,p^{++})=\text{thunk} (\lambda a_1.f(a_1,p^{+})\; p^*)=\text{thunk} (\lambda a_2.\text{thunk} (\lambda a_1.f(a_1,a_2)\; p^*)\; p^*)$.

\section{Higher-order Generalization Bounds}

We start by introducing a general class of deep h-o stochastic classifiers in the type system above, before offering our definition of a h-o generalization bound in this setting.  For convenience, we assume a binary classification setting, although the generalization of our framework to regression and multi-class classification is straightforward.

\textbf{Stochastic classifier models.}  We assume we have input and output types $X$ and $Y$, where for classification $Y=\{0,1\}$.  Further, let $Z$ represent fixed-precision positive and negative reals.  We use the fixed notation $\text{N}(.;\mu,\Sigma)$ to represent a multivariate normal (belonging to type $Z^n \rightarrow I$), and $\text{NN}_{T_1,T_2}(.;\theta)$ to represent a neural network with parameters $\theta$ (belonging to function type $T_1\rightarrow T_2$ for some types $T_1,T_2$).  We then define a hierarchy of types: $F_0=(X\rightarrow Y)$, $F_1=F'_0=(X\rightarrow Y)\rightarrow I$, $F_2=F''_0$, and so on.  Here, $F_0$ is the type of deterministic classifiers between $X$ and $Y$; $F_1$ represents distributions over $F_0$, corresponding to stochastic classifiers; and $F_2$ represents distributions over $F_1$, which may be interpreted as a h-o stochastic classifier (which will be used to represent hyper-priors/posteriors in a meta-learning setting).  We can specify flexible models at all these levels via the following probabilistic programs, $f_0:F_0$, $f_1:F_1$, $f_2:F_2$:
\begin{eqnarray}\label{eq:stoch_class}
f_0 &=& \text{NN}_{X,Y}(.;\theta_0) \nonumber \\
f_1 &=& \text{NN}_{X,Y}(.;\text{NN}_{Z^d,\Theta_0}(z_1^+;\theta_1)+e_1^+) \nonumber \\
f_2 &=& \text{NN}_{X,Y}(.;\text{NN}_{Z^d,\Theta_0}(z_1^{++};\text{NN}_{Z^d,\Theta_1}(z_2^+,\theta_2)+e_2^+)+e_1^{++}) 
\end{eqnarray}
Here, $\Theta_0,\Theta_1$ are the parameter spaces (types)  for $\theta_0,\theta_1$, $z_1, z_2 =\text{N}(.;\mathbf{0}_d,\mathbf{I}_d)$ are standard normal latent variables (where $d$ is the dimensionality of the latent space), and $e_1 =\text{N}(.;\mathbf{0}_{|\Theta_0|},\sigma\mathbf{I}_{|\Theta_0|})$ is a noise term (similarly for $e_2$, substituting $\Theta_1$ for $\Theta_0$).

\textbf{Generalization bounds.}  In the setting above, a generalization bound may be defined as a function $\phi: F_1 \times X^N \rightarrow Z$, which takes a stochastic classifier $f_1:F_1$ and a sample of $S$ size $N$ ($S:X^N$) and returns a real value.  Implicitly, the bound is also paired with an associated distribution over input and output types, $D:(X\times Y)'$, and we additionally require that, with probability $(1-\delta)$, the Gibbs Risk $R$ of $f_1$ is less than the bound when applied to a sample drawn from $D$; hence $P(\text{sample}(f_1)(x)\neq y)\leq_{(1-\delta)}\phi(f_1,S)$.  For convenience, we label the type of generalization bounds as $\Phi$.  A h-o generalization bound may then be defined as a h-o stochastic function, which returns a member of $\Phi$ subject to certain conditions:

\textbf{Definition 1} (Higher-order Generalization Bound). \textit{Using the notation defined above, a {\em higher-order generalization bound} is (a) a stochastic function $\phi_{ho}$ with type $A_1 \times A_2 \times ... \times A_n \rightarrow \Phi$, along with (b) a set of {\em independence assumptions} of the form $a \indep b$, where $a\in\{a_1,...,a_n\}$ and $b\in\{f_1,S\}$.  Additionally, we require that, for $a_1:A_1,...,a_n:A_n$, $f_1:F_1$ and $S\sim D$, we have $R(f_1)=P(\text{\em sample}(f_1)(x)\neq y)\leq_{(1-\delta)}(\phi_{ho}(a_1,...,a_n))(f_1,S)$, assuming the independence assumptions in (b) are met.}

We provide below examples two classes of h-o generalization bound, transfer-/meta-learning variational bounds and learned model complexity bounds, both within a PAC-Bayes setting, and discuss how each can be used to provide general training objectives for DPP optimization.

\subsection{Transfer-/Meta-learning Variational Bounds}\label{sec:3_1}

\textbf{Transfer-Learning.} We begin by stating a basic form of the PAC-Bayes bound from [1], in the general DPP setting:
\begin{eqnarray}\label{eq:pac_bayes1}
\phi^1(f_1^\rho,S) &=& R(f_1^\rho,S) + (1/\lambda)[\text{KL}(f_1^\rho,f_1^\pi)+\log(1/\delta)+(\lambda^2/N)] 
\end{eqnarray}
Here, $f_1^\pi$ and $f_1^\rho$ denote the PAC-Bayes prior and posterior respectively, $R(f_1^\rho,S)=P_S(\text{sample}(f_1)(x)\neq y)$ is the Gibbs risk on $S$, $\lambda$ controls the tightness of the bound, and the remaining notation is as defined above.  As proposed in [3,14], a data-dependent prior may be used in Eq. \ref{eq:pac_bayes1}, which is learned on hold-out data $S'$.  The hold-out data may be sampled from the same distribution $D$ as the classifier $f_1$ is tested on, or a related distribution, $D'$; we refer to the latter case as a transfer-learning setting.  Assuming we have an algorithm for training the prior, $\mathcal{A}:(X\times Y)^N\rightarrow F_1$, we may express the transfer-learning bound as a h-o bound: $\phi^{\text{TL}}(S')=\lambda (f_1^\rho,S).\phi^1(f_1^\rho,S;f_1^\pi=\mathcal{A}(S'))$.  Here, we require the independence assumption $S' \indep S$ for part (b) of Def. 1.  Note however that $S' \indep f_1$ is not required; as discussed in [14], $f_1$ may depend on the combined dataset, $[S' S]$.

We would like to learn bounds of the form $\phi^{\text{TR}}(a,S')$ for flexible DPPs of the kind in Eq. \ref{eq:stoch_class} (unlike the restricted forms of distribution and classifier used in [3,14]).  However, the KL term between two DPPs is typically intractable to evaluate.  We thus derive a modified variational bound (using techniques from [12]), which upper-bounds Eq. \ref{eq:pac_bayes1} (and thus bounds the expected risk), while being tractable to optimize:

\textbf{Theorem 1} (Variational Transfer-Learning Bound). \textit{Using the notation above, with variational distributions represented by DPPs, $r_1:(X,Y)\rightarrow (Z^d)'$ and $r_2:\Theta_0\rightarrow (Z^d)'$, the following forms a valid h-o generalization bound, under the condition $S' \indep S$:}
\begin{eqnarray}\label{eq:pac_bayes1a}
\phi^{\text{TL}}_a(S')&=&\lambda (f_1^\rho,S).\min_{r_1,r_2}\phi^1_a(f_1^\rho,S;r_1,r_2,f_1^\pi=\mathcal{A}(S')) \nonumber\\
\phi^1_a(f_1^\rho,S;r_1,r_2) &=& -\mathbb{E}_{S,r_1(\gamma|x,y)}[\log(f_1^\rho(y|x,\gamma)]+ \mathbb{E}_S[\text{KL}(r_1(\gamma|x,y),z_1)] + \nonumber\\
&&  (1/\lambda)[ \mathbb{E}_{z_1(\gamma)f_1^\rho(\theta_0|\gamma)}[\log z_1(\gamma)+\log f_1^\rho(\theta_0|\gamma) - \log r_2(\gamma|\theta_0)]  - \nonumber\\
&&  \mathbb{E}_{f_1^\rho}[\log(f_1^\pi(\theta_0))] +\log(1/\delta)+(\lambda^2/N)],
\end{eqnarray}
\textit{where we write $f_1(\theta_0)$ for $f_1(f_0=\text{NN}_{X,Y}(.;\theta_0))$; $f_1(.|\gamma)$ for $f_1[\gamma/z_1^+]$, and  $f_1^\rho(y|x,\gamma)=P(\text{sample}(f_1^\rho(.|\gamma))(x)=y)$.  Further, we have that $\phi^{\text{TL}}_a(S')(f_1^\rho,S)\geq \phi^{\text{TL}}(S')(f_1^\rho,S)\geq_{(1-\delta)} R(f_1)$ under the same assumptions.}

{\em Proof.}  We first note that the higher-order bound holds true if we substitute $\phi^1(f_1^\rho,S)$ for $\phi^1_a(f_1^\rho,S;r_1,r_2)$, where $\phi^1(f_1^\rho,S)$ is the PAC-Bayes bound from [1]:
\begin{eqnarray}\label{eq:pac_bayes1b}
\phi^1(f_1^\rho,S) &=& R(f_1^\rho,S) + (1/\lambda)[\text{KL}(f_1^\rho,f_1^\pi)+\log(1/\delta)+(\lambda^2/N)] 
\end{eqnarray}
This follows, since the algorithm used to set the prior in Eq. \ref{eq:pac_bayes1a} is applied to $S'$, and we have by assumption that $S' \indep S$.  We can re-express Eq. \ref{eq:pac_bayes1b} by splitting the KL-term:
\begin{eqnarray}\label{eq:pac_bayes1c}
\phi^1(f_1^\rho,S) &=& R(f_1^\rho,S) + (1/\lambda)[-\mathbb{E}_{f_1^\rho(f_0)}[\log(f_1^\pi(f_0))] - \mathbb{H}(f_1^\rho(f_0))+\log(1/\delta)+(\lambda^2/N)],\nonumber\\
\end{eqnarray}
where $\mathbb{H}(.)$ is the Shannon entropy.  We then note that we can upper-bound the risk $R(f_1^\rho,S)$ by the negative log-likelihood, which in turn can be upper-bounded by the negative-ELBO, introducing the variational conditional distribution $r_1:(X,Y)\rightarrow (Z^d)$:
\begin{eqnarray}\label{eq:pac_bayes1d}
R(f_1^\rho) = \mathbb{E}_{f_1(f_0),S}[f_0(x)\neq y] &\leq& -\mathbb{E}_S[\log(\mathbb{E}_{f_1(f_0)}[f_0(x)=y])] \nonumber\\
&\leq& - \mathbb{E}_{S,r_1(\gamma|x,y)}[\log(f_1^\rho(y|x,\gamma)]+ \mathbb{E}_S[\text{KL}(r_1(\gamma|x,y),z_1)]. \nonumber\\
\end{eqnarray}
Further, we have the following lower-bound on the entropy introduced in [16]: $\mathbb{H}(q(x))\geq -\mathbb{E}_{q(x,\gamma)}[\log q(\gamma)+\log q(x|\gamma) - \log r(\gamma|x)]$.  This can be used to upper-bound the negative entropy term in  Eq. \ref{eq:pac_bayes1b}, introducing the variational distribution $r_2:\Theta_0\rightarrow (Z^d)'$: 
\begin{eqnarray}\label{eq:pac_bayes1e}
- \mathbb{H}(f_1^\rho(f_0)) \leq \mathbb{E}_{z_1(\gamma)f_1^\rho(\theta_0|\gamma)}[\log z_1(\gamma)+\log f_1^\rho(\theta_0|\gamma) - \log r_2(\gamma|\theta_0)].
\end{eqnarray}
Substituting the upper-bounds in Eqs. \ref{eq:pac_bayes1d} and \ref{eq:pac_bayes1e} into Eq. \ref{eq:pac_bayes1b} yields $\phi^1_a$ in Eq. \ref{eq:pac_bayes1a}, and hence we have $\phi^{\text{TL}}_a(S')(f_1^\rho,S)\geq \phi^{\text{TL}}(S')(f_1^\rho,S)\geq_{(1-\delta)} R(f_1)$.
\begin{flushright}
$\square$
\end{flushright}

\textbf{Meta-Learning.} For the case of meta-learning, [4] introduce a bound, which can be expressed in our notation as:
\begin{eqnarray}\label{eq:meta1}
&&\phi^2(f_2^\rho, f_1^{\rho,1}, f_1^{\rho,2}...f_1^{\rho,M}) = \mathbb{E}_{t}[R(f_1^{\rho,t}) + ((\text{KL}(f_2^\rho,f_2^\pi)+\text{KL}(f_1^{\rho,t},(f_2^\pi)^*)+a)/b)^{1/2}] \nonumber \\
&&\quad\quad\quad\quad\quad\quad\quad\quad\quad\quad\;\;+((\text{KL}(f_2^\rho, f_2^\pi)+c)/d)^{1/2}
\end{eqnarray}
where $f_2^\pi$ and $f_2^\rho$ denote a hyper-prior and hyper-posterior respectively (belonging to the type $F_2$, as in Eq. \ref{eq:stoch_class}),   $M$, and $N_{t}$ are the tasks, and training examples for task $t$ respectively, $\mathbb{E}_{t}[.]$ denotes the average as $t$ ranges over tasks, $a=\log(2MN_m/\delta)$, $b=2(N_m-1)$, $c=\log(2M/\delta)$ and $d=2(M-1)$.  We may use Eq. \ref{eq:meta1} to define a meta-learning h-o generalization bound:
\begin{eqnarray}\label{eq:meta2}
\phi^{\text{ML}}(S_1,...,S_M)&=&\lambda (f_1^\rho,S_{M+1}).\phi^1_b(f_1^{\rho,M+1},S_{M+1};f_1^\pi=\text{sample}(\mathcal{A}(S_1,...,S_M))) \nonumber\\
\mathcal{A}(S_1,...,S_M)&=&\text{argmin}_{f_2^\rho}\min_{f_1^{\rho,1...M}} \phi^2(f_2^\rho, f_1^{\rho,1}, f_1^{\rho,2}...f_1^{\rho,M}),
\end{eqnarray}
where we require that the samples from the training tasks $(S_1,...,S_M)$ are independent of the test task $S_{M+1}$, and $\phi^1_b$ is defined in Appendix \ref{app:B}.  Additionally, Eq. \ref{eq:meta2} provides a bound on the transfer error, i.e. the expected error on a new task: $\min_{f_1}\phi^{\text{ML}}(S_1,...,S_M)(f_1,S_{M+1})\leq_{(1-\delta)}\min_{f_1^{\rho,1...M}}\phi^2(\mathcal{A}(S_1,...,S_M),$ $f_1^{\rho,1...M})$ (see [4], Theorem 2).  Again, we would like to learn bounds of the form $\phi^{\text{ML}}$ for flexible DPPs, without the restrictions on distributions used in [4]; for this purpose, a modified variational bound can be derived $\phi^{\text{ML}}_a$ analogously to Theorem 1 for tractable optimization, which we state in Appendix \ref{app:B}.  Finally, we note that Eq. \ref{eq:meta2} can be simplified, following [25], by splitting the samples $S_{1...M+1}$ into training and testing partitions, and learning/fixing a function $V:F_2\times(X\times Y)^N\rightarrow F_1$ to generate the task-specific classifiers using the training sets.  The task priors and posteriors are then set to $V(f^{\pi}_2,S_t^{\text{train}})$ and $V(f^{\rho}_2,S_t^{\text{train}})$ resp., causing the $\text{KL}(f_1^{\rho,t},(f_2^\pi)^*)$ terms to vanish in Eq. \ref{eq:meta1} [25], and $S_{M+1}^{\text{train}}$ is treated as  a further input to the h-o bound.  Alternatively, $V$ may use summary features of the task-samples without creating a train/test partition, and a differential privacy penalty added to the bound as used in the single-task setting in [7].

\subsection{Second-order Complexity Bounds}

The bounds considered in Sec. \ref{sec:3_1} allow for a prior over classifiers to be trained in the case of transfer-learning, or a hyper-prior in the case of meta-learning, assuming the prior is trained on separate data $S'$ from that used to evaluate the bound $S$.  This can be viewed as a form of `learned complexity', since the KL-divergence terms in the bounds outlined penalize the divergence between the posterior and the trained prior, rather than one of generic form, such as a Gaussian or Minimum Description Length (MDL) [26] prior.  However, a more direct way to introduce a learned complexity term is simply to train an additional model $g$ to predict the generalization error.  In this section, we introduce a class of higher-order bounds based on this principle (which we call `second-order complexity' bounds).  These bounds take as input a `generalization predictor' $g_1$, and output a bound.

To simplify the analysis, we treat generalization prediction as a classification task (the regression case is discussed below).   Hence, we introduce the type $G_0=(F_0\rightarrow\{0,1\})$ for a deterministic generalization classifier $g_0:G_0$, and $G_1=G_0'$ for a stochastic classifier $g_1:G_1$.  For a given threshold $\tau$, along with a base classifier of interest $f_0$, $g_1$ will be trained to predict whether the generalization error of $f_0$ exceeds $\tau$,  i.e. $(R(f_0)-R(f_0,S))>\tau$.  The risk of $g_1$ applied to a stochastic base classifier $f_1:F_1$ for threshold $\tau$, can thus be expressed as the risk that $g_1$ incorrectly predicts the generalization error of classifier $f_0$ sampled according to $f_1$:
\begin{eqnarray}\label{eq:cmplx1}
R^{\tau}_{f_1}(g_1)=\sum_{f_0,g_0} f_1(f_0)g_1(g_0)\left[g_0(f_0)\neq[(R(f_0)-R(f_0,S))>\tau]\right].
\end{eqnarray}
Further, we define $P_1(g_1,f_1)$ as the probability that $g_1$ outputs 1 under $f_0$: $P_1(g_1,f_1)=\sum_{f_0,g_0} f_1(f_0)g_1(g_0)\cdot$ $[g_0(f_0)=1]$.  With these definitions, and letting $I^{\tau}_{f_0}(S_1,S_2)=[(R(f_0,S_1)-R(f_0,S_2))>\tau]$ and $S'=\{S'_1,...,S'_{N'}\}$ be a set of $N'$ auxiliary datasets (of arbitrary size) sampled from $D$, we have the following h-o bound:

\textbf{Theorem 2} (Second-order Complexity Bound). \textit{Using the notation above, the following forms a h-o generalization bound, under the condition  $S' \indep (f_1,S)$:}
\begin{eqnarray}\label{eq:cmplx3}
\phi^{\text{2o-cplx}}(g,\tau,S')&=&\lambda (f_1^\rho,S).(R(f_1,S)+\epsilon(f_1,g_1,\tau)) \nonumber\\
\epsilon(f_1,g_1,\tau)&=&\tau + (R^{\tau}_{f_1}(g_1,S_A) + \eta(g_1) + P_1(g_1,f_1))(1-\tau) \nonumber\\
\eta(g_1)&=&\frac{1}{\lambda}\left(\text{KL}(g_1,\pi_1)+\log\left(\frac{1}{\delta}\right)+\left(\frac{\lambda^2}{N'}\right)\right),
\end{eqnarray}
\textit{where $S_A$ is an auxiliary sample, formed by sampling $N'$ values of $f_0$ according to $f_1$, i.e. $\{f_0^{(m)}|m=1...N'\}$, and creating the pairs $(f_0^{(m)},I^{\tau}_{f_0}(S'_m,S))$.  Further, $\pi_1$ is a fixed prior on $g_1$.}

{\em Proof.} From the above, we have:
\begin{eqnarray}\label{eq:cmplx11}
R^{\tau}_{f_1}(g_0)&=& \sum_{f_0} f_1(f_0)\left[g_0(f_0)\neq[(R(f_0)-R(f_0,S))>\tau]\right], \nonumber\\
R^{\tau}_{f_1}(g_1)&=&\sum_{f_0,g_0} f_1(f_0)g_1(g_0)\left[g_0(f_0)\neq[(R(f_0)-R(f_0,S))>\tau]\right], \nonumber\\
R(f_1,S)&=&\frac{1}{|S|}\sum_{f_0,(x,y)\in S} f_1(f_0)\left[f_0(x)\neq y\right], 
\end{eqnarray}
and
\begin{eqnarray}\label{eq:cmplx2}
P_1(g_0,f_1)=\sum_{f_0} f_1(f_0)[g_0(f_0)=1], \nonumber\\
P_1(g_1,f_1)=\sum_{f_0,g_0} f_1(f_0)g_1(g_0)[g_0(f_0)=1], 
\end{eqnarray}
while letting $P_0(g_0,f_1)=1-P_1(g_0,f_1)$, $P_0(g_1,f_1)=1-P_1(g_1,f_1)$.  Next, we observe that the following holds with probability 1:
\begin{eqnarray}\label{eq:cmplx44}
R(f_1) &\leq & R(f_1,S) + \epsilon'(f_1,g_1,\tau) \nonumber\\
\epsilon'(f_1,g_1,\tau)&=&\tau + (R^{\tau}_{f_1}(g_1) + P_1(g_1,f_1))(1-\tau).
\end{eqnarray}
We can demonstrate Eq. \ref{eq:cmplx44} by the following argument.  Observe that, for a given $g_0$, it will classify $f_0$ as having a generalization error less than $\tau$ with probability $P_0(g_0,f_1)$.  However, since its risk of misclassification is $R^{\tau}_{f_1}(g_0)$, we can lower-bound the true 0 outputs (true negatives) by $P_0(g_0,f_1)-R^{\tau}_{f_1}(g_0)$.  By definition, the generalization error of these true negatives is less that $\tau$, and the generalization error in all other cases cannot be more than 1.  Hence, taking a weighted average, a bound on the generalization error for a given $g_0$ can be written as:
\begin{eqnarray}\label{eq:cmplx5}
R(f_1) &\leq & R(f_1,S) + \epsilon'(f_1,g_0,\tau) \nonumber\\
\epsilon'(f_1,g_0,\tau)&=& (P_0(g_0,f_1)-R^{\tau}_{f_1}(g_0))\cdot\tau + (1-P_0(g_0,f_1)+R^{\tau}_{f_1}(g_0))\cdot 1 \nonumber\\
&=& (1-P_1(g_0,f_1)-R^{\tau}_{f_1}(g_0))\cdot\tau + (P_1(g_0,f_1)+R^{\tau}_{f_1}(g_0))\cdot 1 \nonumber\\
&=& \tau +(R^{\tau}_{f_1}(g_0) + P_1(g_0,f_1))(1-\tau).
\end{eqnarray}
Eq. \ref{eq:cmplx44} then follows by taking the expectation of both sides on Eq. \ref{eq:cmplx5} across $g_1$ (i.e. $g_0\sim g_1$) (note that if $P_0(g_0,f_1)-R^{\tau}_{f_1}(g_0)$ is less than 0, the bound is greater than $1$, and hence is valid vacuously).

Next, we wish to replace $R^{\tau}_{f_1}(g_1)$ with an empirical estimate $R^{\tau}_{f_1}(g_1,S_A)$ as in the theorem, where $S_A$ is the auxiliary sample discussed in Sec. 3.2.  To do so, we first consider the risk of $g_1$ not with respect to predicting the true generalization error (Eq. \ref{eq:cmplx11}), but rather on a further sample, $S^{\dagger}$ of size $M$.  We write this:
\begin{eqnarray}\label{eq:cmplx6}
R^{\tau}_{f_1}(g_1,S^{\dagger})=\sum_{f_0,g_0} f_1(f_0)g_1(g_0)\left[g_0(f_0)\neq[(R(f_0,S^{\dagger})-R(f_0,S))>\tau]\right],
\end{eqnarray}
We can then apply the PAC Bayes bound from [1] to the empirical estimate of the risk of $g_1$ on the auxiliary sample, where we assume each of the auxiliary datasets $S'_1,S'_2,...$ has size $M$:
\begin{eqnarray}\label{eq:cmplx7}
\mathbb{E}[R^{\tau}_{f_1}(g_1,S^{\dagger})] \leq_{\delta} R^{\tau}_{f_1}(g_1,S_A) + (1/\lambda)[\text{KL}(g_1,\pi_1)+\log(1/\delta)+(\lambda^2/N')] 
\end{eqnarray}
where $\pi_1$ is an arbitrary prior.  We note that Eq. \ref{eq:cmplx7} requires the assumption (from the theorem) that $S' \indep f_1$: This is because the true risk $R^{\tau}_{f_1}(g_1,S^{\dagger})$ is defined over the product distribution of $D$ (the base-level distribution over $(X,Y)$) and $f_1$; hence $S_A$ will be a sample from the same distribution iff  $S' \indep f_1$.  We then observe that, using Eq. \ref{eq:cmplx7}:
\begin{eqnarray}\label{eq:cmplx8}
R^{\tau}_{f_1}(g_1) = \mathbb{E}[R^{\tau}_{f_1}(g_1,S^{\dagger})] \leq_{\delta} R^{\tau}_{f_1}(g_1,S_A) + \eta(g_1),
\end{eqnarray}
where $\eta(g_1)$ is as in Eq. \ref{eq:cmplx3}.  Finally,  substituting the bound for $R^{\tau}_{f_1}(g_1)$ in Eq. \ref{eq:cmplx8} into Eq. \ref{eq:cmplx44}, results in the h-o bound given in Eq. \ref{eq:cmplx3}.  
\begin{flushright}
$\square$
\end{flushright}

We note that the bound in Theorem 2 contains the term $P_1(g_1,f_1)$, which requires an estimate of the probability the stochastic classifier $g_1$ will return 1 under inputs from $f_1$ (Eq. \ref{eq:cmplx2}).  However, this quantity does not depend on external data (i.e. either $S$ or $S'$), and hence it can be made arbitrarily precise by repeatedly drawing samples $g_0$ and $f_0$ from $g_1$ and $f_1$ resp. and observing $g_0(f_0)$, allowing the bound to be evaluated to arbitrary accuracy.  Further, the bound may be used during training in a number of distinct ways.  Most directly, a stochastic classifier $f_1$ may be trained on $S$ and then fixed;  Eq. \ref{eq:cmplx3} may then be optimized, leading to the bound $R(f_1)\leq_{(1-\delta)}\phi^{\text{2o-cplx}}(g_1^*,\tau^*,S')$, where $(g_1^*,\tau^*)=\text{argmin}_{(g_1,\tau)}\phi^{\text{2o-cplx}}(g,\tau,S')(f_1,S)$. Alternatively, a (small) set of stochastic classifiers may be considered, $\mathcal{F}$ (for instance, those generated over an optimization path when training $f_1$ on $S$); the bound may then be optimized to pick the final classifier: $f_1^*=\text{argmin}_{f_1\in\mathcal{F}}\min_{(g_1,\tau)}\phi^{\text{2o-cplx}}(g,\tau,S')(f_1,S)$, after applying a union bound.  Another possibility is to directly optimize the bound over $(f_1,g_1,\tau)$, while applying a differential privacy transformation to $f_1$ when calculating $\epsilon(f_1,g_1,\tau)$, to approximately enforce $S' \indep f_1$.  This results in the following optimization problem:
\begin{eqnarray}\label{eq:cmplx9}
(f_1^*,g_1^*,\tau^*) = \text{argmin}_{(g_1,\tau)}\phi^{\text{2o-cplx}}(g,\tau,S')(h(f_1),S),
\end{eqnarray}
where $h(.)$ is a privacy preserving transformation.  A generic form for $h(.)$ is given below:
\begin{eqnarray}\label{eq:cmplx10}
h(f_1)=\lambda f_0.\left(\frac{\exp(\beta\log f_1(f_0))}{\sum_{f_0}\exp(\beta\log f_1(f_0))}\right),
\end{eqnarray}
which increases the entropy of $f_1$ according to the `temperature' $\beta$.  The bound in Th. 2 may then be modified by incorporating an additional differential privacy term, as in [7].  Additionally, we note that while Th. 2 uses a generalization error classifier, $g_1$, equally we may consider the case of a  2-o complexity bound based on  a generalization error regressor $g^{\text{reg}}_1:(F_0\rightarrow \mathbb{R})'$.  Here, $g_1^{\text{reg}}$ would be trained directly to predict $(R(f_0)-R(f_0,S))$.  Investigation of the analogous bound for optimizing $(f_1,g^{\text{reg}}_1)$ is left to future work.

Finally, we may form a meta-learning analogue of $\phi^{\text{2o-cplx}}$ (see Appendix \ref{app:C}):

\textbf{Theorem 3} (Second-order Complexity ML-Bound). \textit{With notation as in Theorem 2, and assuming $(S_1,...S_M,S'_1,...S'_{M+1})\indep  (f_1,S_{M+1})$, we have the h-o bound:}
\begin{eqnarray}\label{eq:cmplx4}
\phi^{\text{2o-cplx-ML}}(S_{1:M},\mathcal{A}_f)&=&\lambda (f_1,S_{M+1}).\phi^{\text{2o-cplx}}(g=\text{sample}(\mathcal{A}_g(S_{1:M})),\tau,S'_{M+1})(f_1,S_{M+1})\nonumber\\
\mathcal{A}_g(S_{1:M})&=&\text{argmin}_{g_2} \mathbb{E}_{t}[\phi^{\text{2o-cplx}}(g=\text{sample}(g_2),\tau,S'_t)(\mathcal{A}_f(S_{t}),S_{t})] + \eta(g_2)\nonumber\\
\eta(g_2)&=&\frac{1}{\lambda}\left(\text{KL}(g_2,\pi_2)+\log\left(\frac{1}{\delta}\right)+\left(\frac{\lambda^2}{M}\right)\right),
\end{eqnarray}
\textit{where $g_2:G_2=G'_1$, $\mathcal{A}_f:(X\times Y)\rightarrow F_1$, and each task $t$ has its own auxiliary data samples, $S'_{t,1:N'_t}$.  Further, a bound on the transfer error is provided by $\mathbb{E}_{t}[\phi^{\text{2o-cplx}}(g=\text{sample}(g^*_2),\tau,S'_t)(\mathcal{A}_f(S_t),S_{t})] + \eta(g_2)$, where $g^*_2=\mathcal{A}_g(S_1,...,S_M)$.}

\section{Results}

\subsection{Synthetic Experiments}

\begin{figure*}[!t]
\centering
\includegraphics[width=4.5in]{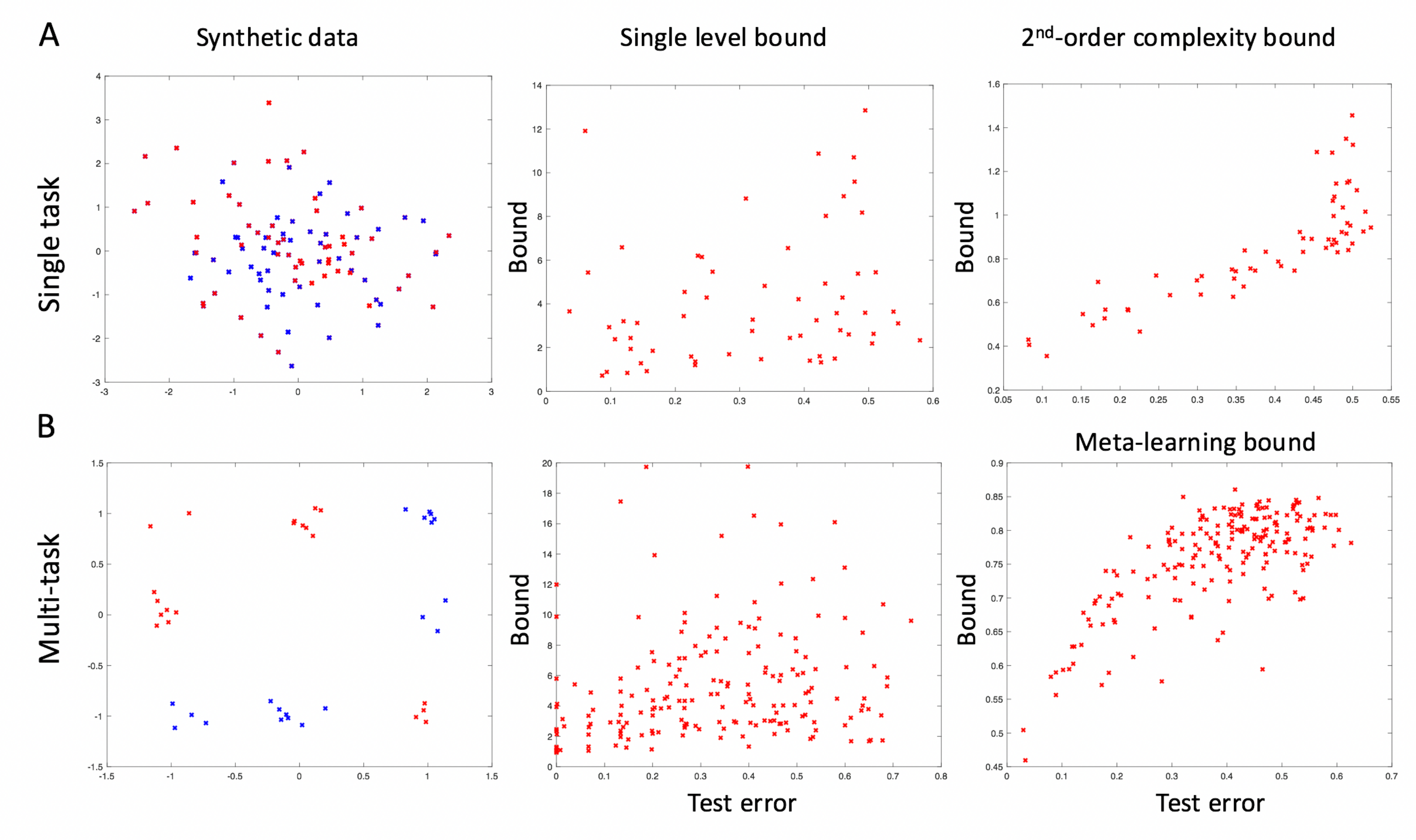}
\caption{Synthetic experiments.  (A) and (B) show example data and results on single- and multi-task synthetic generalization tasks respectively.  The left panel shows example data from a task in each case.  The middle panel compares the test classification error and generalization bound predicted by a single-level variational PAC-Bayes bound ($\phi^1_a$, Eq. \ref{eq:pac_bayes1a}).  The third panel similarly compares test error and bound for the second-order complexity bound ($\phi^{\text{2o-cplx}}$, Eq. \ref{eq:cmplx3}) and DPP meta-learning bound ($\phi^{\text{ML}}$, Eq. \ref{eq:meta2}) for top and bottom respectively.}
\label{fig:fig1}
\end{figure*}

We first design a single-task classification experiment, to compare the performance of the single-level variational PAC-Bayes ($\phi^1_a$, Eq. \ref{eq:pac_bayes1a}) bound against the second-order complexity bound ($\phi^{\text{2o-cplx}}$, Eq. \ref{eq:cmplx3}).  We create 10 synthetic datasets with 100 testing, validation and training points each as illustrated in Fig. \ref{fig:fig1}A.  We use a generative process for the data involving first, sampling 20 `prototype' points, each sampled from a standard 2-d Gaussian, and labeled 0/1 such that (a) the first two prototypes are 0 and 1 resp., (b) prototypes 3-20 are labeled by finding the nearest neighbor from the previous prototypes, and flipping this label with a 0.1 probability.  The train/validation/test data points are then sampled likewise from the 2-d standard Gaussian, and labeled according to their nearest prototype (while balancing each set to contain 50 points from each class).  This process is designed to generate synthetic data with a complex decision boundary, which has structure on multiple scales.  We evaluate the ability of the bounds $\phi^1_a$ and $\phi^{\text{2o-cplx}}$ to predict generalization performance on 60 networks trained on these 10 datasets, where for each dataset we add a varying amount of label noise, corresponding to 0, 20, 40, 60, 80 and 100\% of the labels being flipped.  For  $\phi^1_a$, we learn a stochastic classifier $f_1^\rho$ using Eq. \ref{eq:pac_bayes1a} on the training partition, after pre-training a prior $f_1^\pi$ on the validation partition using the ELBO bound [12].   For $\phi^{\text{2o-cplx}}$, we train $f_1$ directly on the training partition using the ELBO bound, and optimize $\phi^{\text{2o-cplx}}$ for $(g_1,\tau)$ by using the validation partition to construct an auxiliary dataset with $N'=20$ bootstrapped samples.  For the second-order generalization classifier $g$, we use 6 network features as predictors: the $\ell_1$ and $\ell_2$-norms of the weights, the likelihood, log-likelihood and entropy of the stochastic classifier on the training set, and the path-norm [13] of the weights.  In all cases, we use networks with 2 hidden layers of 5 units each, a 2-d latent space, set $\sigma=0.1, \lambda=10, \delta=0.05$.  The results in Fig. \ref{fig:fig1}A show that both bounds are able to predict generalization performance.  The traditional PAC-Bayes bound bound ($\phi^1_a$) achieves a moderate correlation with the test error ($r=0.22$, $p=0.086$), while the second-order complexity bound ($\phi^{\text{2o-cplx}}$) achieves a stronger correlation ($r=0.46$, $p=1.9e-4$).  Further, the second-order complexity bound is shown to carry significant additional information about the test error versus the training and validation error alone ($p=0.03$ and $p=0.02$ resp., 1-tailed ANOVA), while the single-layer bound is only weakly informative ($p>0.1$), suggesting that the $\phi^{\text{2o-cplx}}$ provides a more efficient representation for data-driven complexity than a PAC-Bayesian data-dependent prior (we note that both methods had access to the same training/validation data during optimization).

Next, we design a multi-task synthetic classification experiment, to compare the single-level and meta-learning variational PAC-Bayes bounds ($\phi^{\text{ML}}$, Eq. \ref{eq:meta2}).  Here, we are particularly interested in the extra flexibility afforded by the modified DPP versions of these bounds, in comparison the the restricted forms used in previous work [4,25].  For this purpose, we design a synthetic dataset, having 33 tasks, each being a binary classification problem with 2d input features, where the inputs fall into 8 Gaussian clusters ($\sigma=0.1$) arranged on the corners and mid-points of a square around the origin, as shown in Fig. \ref{fig:fig1}B, with 4 being randomly assigned to classes 0 and 1 on each task. This allows for transfer of information across tasks, since similar decision boundaries may occur in multiple tasks. For each task, we generate 6 datasets with varying levels of noise added (to permit different levels of generalization), flipping 0, 20, 40, 60, 80 and 100\% of the labels, and split the data into training, validation and testing partitions of $N=15$ data-points each. We first learn a stochastic classifier $f_1^\rho$ using Eq. \ref{eq:pac_bayes1a} on the validation partition, after pre-training a prior $f_1^\pi$ on the training partition using the ELBO bound [12].  Fig. \ref{fig:fig1}B plots the test error against the bound, which are significantly correlated ($r=0.2, p=0.008$).  Further, a regression of the test error on the training error and bound show the bound to be moderately informative ($p=0.1$, 1-tailed ANOVA).  We then use the DPP meta-learning bound ($\phi^{\text{ML}}$) to learn classifiers $f_1^\rho$ for each task, while simultaneously fitting a hyper-posterior $f_2^\rho$ to groups of 3 tasks at a time (using the validation sets only).  Fig. \ref{fig:fig1}B shows this approach is able to achieve a better correlation between the bound and test error ($r=0.7, p=2e-30$), and that the bound carries significant additional information about the test error versus the training error alone ($p=0.01$, 1-tailed ANOVA), showing that the meta-learning approach is able to share information between tasks. We compare against the model of [4], in which the priors, and hyper-posterior/prior are restricted to be Gaussian in form, which achieves significantly lower test performance across tasks ($p=0.015$, 1-tailed t-test, $0.53$ vs $0.56$ mean accuracy), showing the flexibility afforded by the DPP formulation to be beneficial.  Network hyper-parameters were set identically to the single-task setting.

%In all cases, we use networks with 2 hidden layers of 5 units each, a 2-d latent space, set $\sigma=0.1, \lambda=10, \delta=0.05$.

\subsection{Modeling psychiatric genomics expression data.}

\begin{figure*}[!t]
\centering
\includegraphics[width=4.5in]{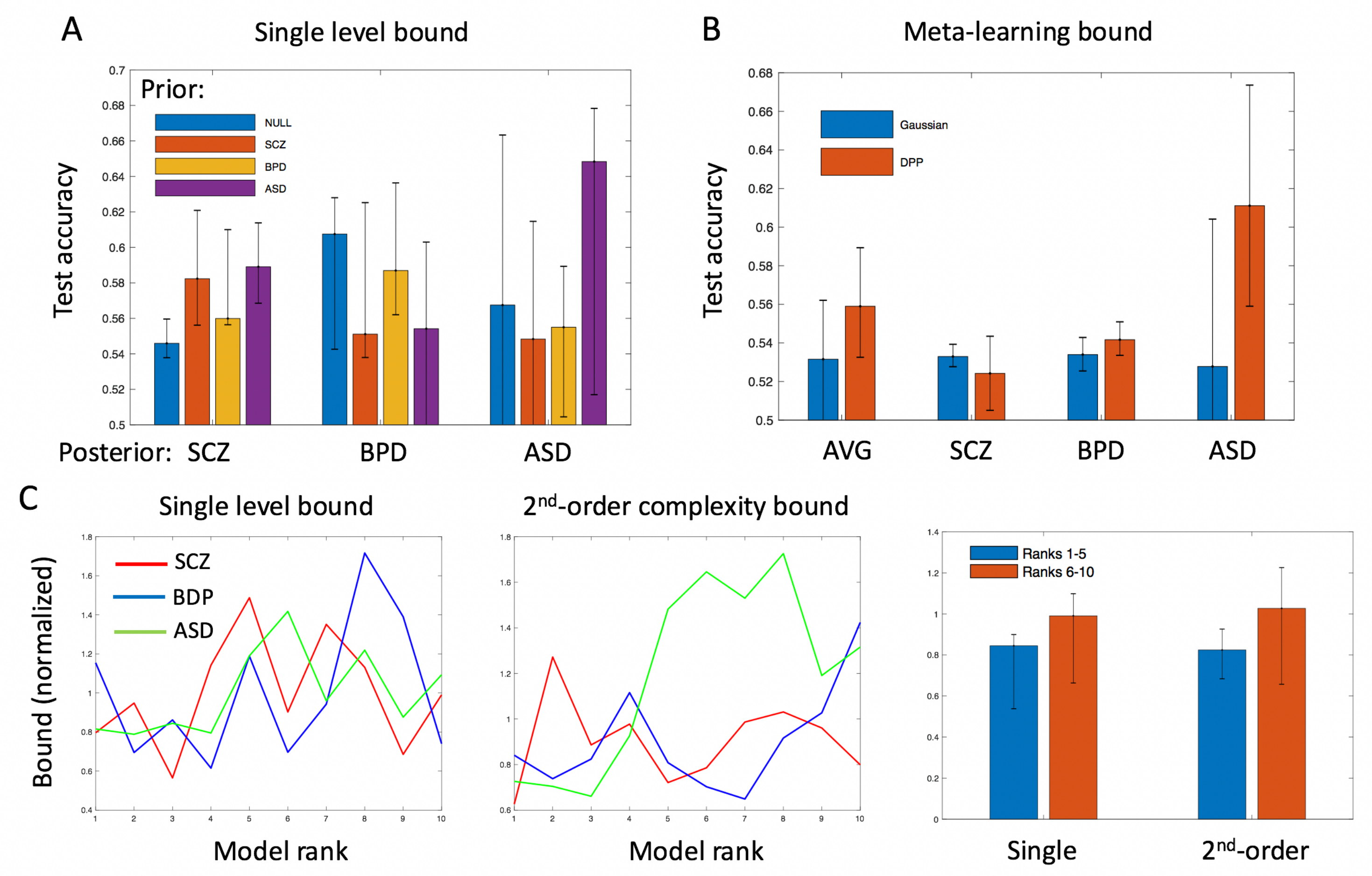}
\caption{Psychiatric genomics expression data.  (A) shows results for transfer-learning on genomics data, where prior and posterior are trained to identify different psychiatric conditions vs controls (Eq. \ref{eq:pac_bayes1}), and (B) compares meta-learning performance on genomics data using a restricted (Gaussian-based, [4]) and full DPP-based model. (C) compares model-selection performance (10 models) using single-level PAC-Bayes ($\phi^1_a$, Eq. \ref{eq:pac_bayes1a}) and second-order complexity bounds ($\phi^{\text{2o-cplx}}$).  Panels 1-2 show normalized bound values (scaled by mean) ordered by true model rank on test data, while panel 3 compares  ranks 1-5 with ranks 6-10 across disorders for each bound.  Error bars show quartiles.}
\label{fig:fig2}
\end{figure*}

We further test our approach on psychiatric genomics data from the PsychENCODE project [23], consisting of gene expression (RNA-Seq) levels from post-mortem prefrontal cortex samples of control, schizophrenia (SCZ), bipolar (BDP) and autistic (ASD) subjects.  We create datasets balanced for cases and controls (and covariates, see [23]) for each disorder, with 710, 188 and 62 subjects respectively, from which we create 10 training, validation and testing partitions (approx. 0.45/0.45/0.1 split).  For each data split, we select the 5 most discriminative genes for each disorder using the training partitions to create a 15-d input space; the network hyper-parameters and bound optimized are identical to the synthetic experiments.  We first test the ability of our approach to perform \textbf{transfer learning}, by training priors $f_1^\pi$ on each of the training partitions (via an ELBO objective), before training a posterior stochastic classifier $f_1^\rho$ using Eq. \ref{eq:pac_bayes1a} on the validation data (via optimizing Eq. \ref{eq:pac_bayes1a}); in doing so, we test all combinations of disorders when learning priors and posteriors.   The results in Fig. \ref{fig:fig2}A show that both SCZ and ASD models are able to use the information in the prior to improve generalization.  The SCZ results are particularly interesting, in that the priors trained on all 3 disorders are able to improve the baseline model; the improvements for the SCZ and BPD priors here are significant ($p=0.006$ and $p=0.026$ respectively, 1-tailed t-test).  In the ASD case, only the ASD prior improves performance, while for BPD, no improvement is gained.  We note that the SCZ dataset is substantially larger than the other disorders', which may affect the results. We also compare against a model with a Gaussian prior, observing lower performance across models ($p=9.9e-3$, 1-tailed t-test, $0.57$ vs $0.59$ mean accuracy). We then test the framework in the \textbf{meta-learning} setting, by optimizing DPP meta-learning bound ($\phi^{\text{ML}}$, Eq. \ref{eq:meta2}) for each of the 10 data-splits on all tasks (SCZ, BPD, ASD classification) simultaneously.  As in the synthetic setting, we also train a model in which the priors, and hyper-posterior/prior are restricted to be Gaussian in form, replicating the setting of [4].  Fig. \ref{fig:fig2}B shows that the DPP meta-learning model is able to achieve better test performance overall ($p=0.13$, 1-tailed t-test), particularly by improving prediction on the BPD and ASD tasks.  We note that, in general, the performance of the models in Fig. \ref{fig:fig2}B is slightly lower than \ref{fig:fig2}A, since we used a limited subset of the data in training the former (56 samples each) in order to balance the data across tasks.  In general, the results of the transfer and meta-learning tasks point to a shared etiology of psychiatric conditions, as has been highlighted recently [2,9].  Finally, we also compare the ability of $\phi^1_a$ and $\phi^{\text{2o-cplx}}$ to predict generalization performance across models trained on each of the 10 data-splits for each disorder, hence performing model selection, using the same (non-transfer) setting as for the synthetic data (Fig. \ref{fig:fig1}A, i.e. training and validation are from the same disorder).  Fig. \ref{fig:fig2}C shows how the bound varies with the actual test-set ranking for each disorder.  The traditional (data-dependent) PAC-Bayes bound achieves a moderate correlation ($r=0.30$, $p=0.11$, panel 1), while the second-order complexity bound is notably stronger ($r=0.42$, $p=0.02$, panel 2), suggesting, as in the synthetic results, that the second-order complexity bound provides a more efficient representation for data-driven complexity.  Further, comparing bound values for ranks 1-5 versus 6-10 across disorders reveals a more significant separation for $\phi^{\text{2o-cplx}}$  than $\phi^1_a$ ($p=0.023$ versus $p=0.077$ respectively, see Fig. \ref{fig:fig2}C panel 3).

\section{Discussion}

We have introduced a framework for deriving higher-order generalization bounds in a DPP context, and have shown that these lead to efficient variational objectives for training DPPs, as well as allowing novel generalization bounds to be derived.  Particularly, we show that a second-order complexity bound we introduce outperforms traditional PAC-Bayes bounds in predicting generalization and model selection on synthetic and genomics tasks.  Our results suggest a number of future directions.  First, as discussed, it is straightforward to include task-based features in the DPP framework to conditionalize the higher-order bounds for transfer- and meta-learning settings.  Further, we note that second-order complexity bound $\phi^{\text{2o-cplx}}$ may naturally be modified by incorporating differential-privacy constraints (following [7]) to mitigate the independence requirements of the bound, and allow joint training of $f_1$ and $g_1$; additionally, the generalization classifier $g_1$ may itself be a more complex program, such as a data-dependent compression algorithm, hence forming a second-order analogue of the MDL bound in [26].  Potentially, exploring forms of $\phi^{\text{2o-cplx}}$ incorporating differential-privacy and MDL priors offers the possibility of deriving tight absolute bounds on generalization as in [26] (here, we have focused on looser bounds as training objectives, and their empirical correlation with test-set generalization).  Finally, we note that while we have assumed a discrete setting for formalizing DPPs, our framework may naturally be lifted to a continuous setting, for instance with a denotational semantics based on Quasi-Borel spaces [11,18], while incorporating the distinction between distributional and thunked types from [20].  More generally, analogues of our bounds may be formulated in distinct probabilistic programming paradigms (for instance, using a factor-graph semantics as in [5]), motivating novel training algorithms based on principled objectives.

% which is trained to compress networks relevant to the task in question

\section*{References}

%\medskip

%\small

[1] Alquier, P., Ridgway, J., \& Chopin, N. (2016). On the properties of variational approximations of Gibbs posteriors. {\em Journal of Machine Learning Research}, 17(1), 8374-8414.

\noindent[2] Anttila, V., Bulik-Sullivan, B., Finucane, H. K., Walters, R. K., Bras, J., Duncan, L., ... \& Neale, B. (2018). Analysis of shared heritability in common disorders of the brain. {\em Science}, 360(6395), eaap8757.

\noindent[3] Ambroladze, A., Parrado-Hernández, E., \& Shawe-taylor, J. S. (2007). Tighter PAC-Bayes bounds. In {\em Advances in neural information processing systems} (pp. 9-16).

\noindent[4] Amit, R. and Meir, R., 2017. Meta-learning by adjusting priors based on extended PAC-Bayes theory. arXiv preprint arXiv:1711.01244.

\noindent[5] Borgström, J., Gordon, A.D., Greenberg, M., Margetson, J. and Van Gael, J., 2011, March. Measure transformer semantics for Bayesian machine learning. In European symposium on programming (pp. 77-96). Springer, Berlin, Heidelberg.

\noindent[6] Dieng, A. B., Tran, D., Ranganath, R., Paisley, J., \& Blei, D. (2017). Variational Inference via $\chi $ Upper Bound Minimization. In {\em Advances in Neural Information Processing Systems} (pp. 2732-2741).

\noindent[7] Dziugaite, G. K., \& Roy, D. M. (2018). Data-dependent PAC-Bayes priors via differential privacy. In {\em Advances in Neural Information Processing Systems} (pp. 8430-8441).

\noindent[8] Dziugaite, G.K. and Roy, D.M., 2017. Entropy-SGD optimizes the prior of a PAC-Bayes bound: Generalization properties of Entropy-SGD and data-dependent priors. arXiv preprint arXiv:1712.09376.

\noindent[9] Gandal, M. J., Haney, J. R., Parikshak, N. N., Leppa, V., Ramaswami, G., Hartl, C., ... \& Geschwind, D. (2018). Shared molecular neuropathology across major psychiatric disorders parallels polygenic overlap. {\em Science}, 359(6376), 693-697.

\noindent[10] Goodman, N., Mansinghka, V., Roy, D. M., Bonawitz, K., \& Tenenbaum, J. B. (2012). Church: a language for generative models. arXiv preprint arXiv:1206.3255.

\noindent[11] Heunen, C., Kammar, O., Staton, S. and Yang, H., 2017, June. A convenient category for higher-order probability theory. In 2017 32nd Annual ACM/IEEE Symposium on Logic in Computer Science (LICS) (pp. 1-12). IEEE.

\noindent[12] Kingma, D. P., \& Welling, M. (2013). Auto-encoding variational bayes. arXiv preprint arXiv:1312.6114.

\noindent[13] Neyshabur, B., Salakhutdinov, R.R. and Srebro, N., 2015. Path-sgd: Path-normalized optimization in deep neural networks. In Advances in Neural Information Processing Systems (pp. 2422-2430).

\noindent[14] Parrado-Hernández, E., Ambroladze, A., Shawe-Taylor, J., \& Sun, S. (2012). PAC-Bayes bounds with data dependent priors. {\em Journal of Machine Learning Research}, 13(Dec), 3507-3531.

\noindent[15] Pentina, A. and Lampert, C., 2014. A PAC-Bayesian bound for lifelong learning. In {\em International Conference on Machine Learning} (pp. 991-999).

\noindent[16] Ranganath, R., Tran, D., \& Blei, D. (2016, June). Hierarchical variational models. In {\em International Conference on Machine Learning} (pp. 324-333).

\noindent[17] Rivasplata, O., Szepesvari, C., Shawe-Taylor, J. S., Parrado-Hernandez, E., \& Sun, S. (2018). PAC-Bayes bounds for stable algorithms with instance-dependent priors. In {\em Advances in Neural Information Processing Systems} (pp. 9214-9224).

\noindent[18] Ścibior, A., Kammar, O., Vákár, M., Staton, S., Yang, H., Cai, Y., Ostermann, K., Moss, S.K., Heunen, C. and Ghahramani, Z., 2017. Denotational validation of higher-order Bayesian inference. Proceedings of the ACM on Programming Languages, 2(POPL), p.60.

\noindent[19] Sobolev, A. and Vetrov, D., 2019. Importance Weighted Hierarchical Variational Inference. arXiv preprint arXiv:1905.03290.

\noindent[20] Staton, S., Wood, F., Yang, H., Heunen, C. and Kammar, O., 2016, July. Semantics for probabilistic programming: higher-order functions, continuous distributions, and soft constraints. In 2016 31st Annual ACM/IEEE Symposium on Logic in Computer Science (LICS) (pp. 1-10). IEEE.

\noindent[21] Tran, D., Hoffman, M. D., Saurous, R. A., Brevdo, E., Murphy, K., \& Blei, D. M. (2017). Deep probabilistic programming. arXiv preprint arXiv:1701.03757.

\noindent[22] Tran, D., Hoffman, M. W., Moore, D., Suter, C., Vasudevan, S., \& Radul, A. (2018). Simple, distributed, and accelerated probabilistic programming. In {\em Advances in Neural Information Processing Systems} (pp. 7598-7609).

\noindent[23] Wang, D., Liu, S., Warrell, J., Won, H., Shi, X., Navarro, F. C., ... \& Gerstein, M. (2018). Comprehensive functional genomic resource and integrative model for the human brain. {\em Science}, 362(6420), eaat8464.

\noindent[24] Warrell J., \& Gerstein M. (2018) Dependent Type Networks: A Probabilistic Logic via the Curry-Howard Correspondence in a System of Probabilistic Dependent Types. In {\em Uncertainty in Artificial Intelligence, Workshop on Uncertainty in Deep Learning}. \\
\noindent http://www.gatsby.ucl.ac.uk/\textasciitilde balaji/udl-camera-ready/UDL-19.pdf

\noindent[25] Yin, M., Tucker, G., Zhou, M., Levine, S. and Finn, C., 2019. Meta-Learning without Memorization. {\em ICLR}, 2020.

\noindent[26] Zhou, W., Veitch, V., Austern, M., Adams, R. P., \& Orbanz, P. (2018). Non-vacuous generalization bounds at the imagenet scale: a PAC-bayesian compression approach. arXiv preprint arXiv:1804.05862.

\newpage

\section*{Appendices}

\appendix
\section{Stochastic Type System}\label{app:A}

In Sec. 2 of the main paper, we describe a stochastic type system based on the higher-order formal language for probabilistic programs stated in [20], with several key differences which we describe below in detail.  Formally, we use the following syntax for types:
\begin{eqnarray}\label{eq:typeSystem11}
\mathbb{A},\mathbb{B} &::=& \mathbb{R} \;|\; \text{P}(\mathbb{A}) \;|\; 1 \;|\; \mathbb{A}\times \mathbb{B} \;|\; \sum_i \mathbb{A}_i \;|\; \mathbb{A}\rightarrow \mathbb{B}.
\end{eqnarray}
Unlike [20], where $\mathbb{A},\mathbb{B}$ are measurable spaces, we will not use the measure structure on these spaces, and hence they can be arbitrary. $\mathbb{R}$ may be interpreted as the continuous reals, or for convenience a discrete representation of the reals to a fixed level of precision as suggested in the main paper.  Further, the constructor $\text{P}(\mathbb{A})$ is taken to represent not the type of probability measures over $\mathbb{A}$ as in [20], but instead the type of finite normalized mass functions over $\mathbb{A}$, where a {\em mass function} over a set $X$ is defined as in [18], as a function $\mu:X\rightarrow \mathbb{R}_+$ for which there exists a finite set $F\subseteq X$ such that $\mu$ is 0 outside $F$, and a normalized mass function is one that sums to 1 over all (deterministic) values of a type.  Hence, for $p:\text{P}(A)$, we may write $p(a)$ for the mass assigned to $a:A$ by $p$, unlike in [20] where $p$  requires a measurable set $U\subseteq A$ as an argument.  Further, we write $A'$ as a synonym for $\text{P}(A)$.  The other constructions in Eq. \ref{eq:typeSystem11} are standard (the unit type, product, sum and function types).

We follow [20] in distinguishing between deterministic and probabilistic typing judgements, written $\Gamma \vdash_{\text{d}} t:A$ and $\Gamma \vdash_{\text{p}} t:A$ respectively, where $\Gamma=\{a_1:A_1, a_2:A_2,...\}$ is a {\em context} of paired term-type assignments.  As in [20], we include the standard constructors/destructors for sum, product and function types (see [20] Secs. 3 and 6), and the probabilistic constructor for sampling, which from $\Gamma \vdash_{\text{d}} t:\text{P}(A)$ allows us to derive $\Gamma \vdash_{\text{p}} \text{sample}(t):A$.  Like [20], we include primitives in the language for standard functions and distributions, for instance, in our case using $\text{N}(.)$ and $\text{NN}(.)$ to denote normal distributions and neural networks, as in the main paper (noting that, since our system is based on normalized mass functions, $\text{N}(.)$ must be a discretized and bounded approximation to a normal distribution, such as one whose support includes only values with a fixed level of precision within a range determined by the CDF).  In addition, we allow probabilistically typed terms to be assigned to $\text{P}(.)$ types through `thunking': Hence, from $\Gamma \vdash_{\text{p}} t:A$ we can derive $\Gamma \vdash_{\text{d}} \text{thunk}(t):\text{P}(A)$, with the proviso that $t$ reduces probabilistically only to a finite number of terms in $A$ (hence ensuring the thunked expression represents a valid finite normalized mass function; this will be ensured if all probabilistically typed terms are constructed from primitive $\text{sample}(.)$ statements).  As stated in the paper, we then require that the following is true: $(\text{thunk}(a))(a')=P(a\rightarrow_{\beta} a')$, where $\rightarrow_{\beta}$ is probabilistic beta-reduction (discussed below).  Our approach to thunking here differs from [20], where a separate  type constructor is introduced for thunked types ($\text{T}(A)$).  The approach in [20] allows thunking to interact with other features of the language (scoring and normalization, based on the side effects of running $a$) which we do not use; hence to simplify the presentation, we use a compact language in which  the constructor $\text{T}(A)$ is not used.  We note that we do not require a special form of equality for $\text{P}(A)$ types: only terms of $\text{P}(A)$ which reduce to the same normal form are regarded as equal (intensional equality), and hence there may be many representations for the same finite normalized mass function (e.g. thunked and non-thunked expressions, or alternative (non-)thunked expressions) which are semantically equivalent, but intensionally non-equal.  Further, our use of finite normalized mass functions for $\text{P}(.)$ means that this constructor can be applied to function types $A\rightarrow B$ and other probabilistic types; hence we may form $\text{P}(A\rightarrow B)$ and $\text{P}(\text{P}(A))=A''$ (noting that the latter has support over a finite number of normal forms in $\text{P}(A)$, which may include thunked and non-thunked values).  This is unlike [20], where the absence of a measure on $A\rightarrow B$, $\text{P}(A)$ and $\text{T}(A)$ prevents the constructors $\text{P}(A)$ and $\text{T}(A)$ being applied recursively.

Finally, we note that we assume an operational semantics which is equivalent to that outlined in [20] (Secs. 5 and 7) to define the stochastic reduction relation between terms (notated above as probabilistic beta-reduction, $\rightarrow_{\beta}$).  For our system, we restrict the semantics outlined in [20] to finite discrete probability measures (as denoted by our $\text{P}(A)$ type), replacing measurable sets with deterministic values, and the $\text{T}(.)$ and $\text{force}(.)$ constructions with $\text{P}(.)$ and $\text{sample}(.)$ constructions as detailed above.  Following [20], the resulting operational semantics requires that only deterministic values can be substituted into function bodies: hence $(\lambda (x:A).B)(\text{sample}(a))$ must be first reduced to $(\lambda x.B)(a_1)$ (for a particular $a_1$), before being reduced to $B[a_1/x]$, and hence if $x$ appears multiple times in $B$, the occurrences will receive the same value rather than being independently sampled (enforcing {\em memoization} as in [9]).  If independent samples are required, $B$ may be modified so that the occurrences of $x$ are labeled differently, e.g. $x_1,x_2$, or a probabilistic/thunked type is used as input, e.g. $(\lambda (x:A').B[\text{sample}(x)/x])(p)$.  We note that, in Eq. 1 from the main paper, we implicitly required that, for a term $f$ containing multiple sampling statements $p_1^+$, these samples should be subject to memoization during evaluation.  If we require instead that certain sampling statements are tied through memoization and others not, the notation in Eq. 1 from the main paper may be adapted to reflect this; hence we may write $f(p_1^+,p_1^+,p_2^+)$ for a DPP in which the two $p_1$ arguments are subject to memoization, and $f(p_1^{+(a)},p_1^{+(b)},p_2^+)$
where they require independent sampling.  To incorporate this notation, the construction may be modified:
\begin{eqnarray}\label{eq:typeSystem111}
&& f[p_1^+,p_1^{+(a)},p_1^{+(b)},...,p_2^+,p_2^{+(a)},...] = \text{thunk}(\lambda (a_1,a_{1a},a_{1b},...,a_2,a_{2a}). \nonumber \\
&& \quad\quad f^{(-)}[a_1/p_1^+,a_{1a}/p_1^{+(a)},a_{1b}/p_1^{+(b)},...,a_{2}/p_2^+,a_{2a}/p_2^{+(a)},...]\; (p_1^*,p_1^*,p_1^*,...,p_2^*,p_2^*,...))\nonumber \\
\end{eqnarray}

\section{Transfer-/Meta-learning Variational Bounds}\label{app:B}

We provide here further results associated with Sec. 3.1 of the main paper.  We first restate the meta-learning bound from [4] in our framework as a higher-order bound (Eq. 6 in the main paper):
\begin{eqnarray}\label{eq:meta22}
\phi^{\text{ML}}(S_1,...,S_M)&=&\lambda (f_1^\rho,S_{M+1}).\phi^1_b(f_1^{\rho,M+1},S_{M+1};f_2^\pi=\mathcal{A}(S_1,...,S_M)) \nonumber\\
\mathcal{A}(S_1,...,S_M)&=&\text{argmin}_{f_2^\rho}\min_{f_1^{\rho,1...M}} \phi^2(f_2^\rho, f_1^{\rho,1}, f_1^{\rho,2}...f_1^{\rho,M})
\end{eqnarray}
where we define $\phi^1_b$ as:
\begin{eqnarray}\label{eq:meta3}
\phi^1_b(f_1^{\rho},S;f_2^\pi) = R(f_1^\rho,S) + \mathbb{E}_{f_1^\pi}((\text{KL}(f_2^\rho,f_2^\pi)+\text{KL}(f_1^{\rho,t},f_1^\pi=\text{sample}(f_2^\pi))+a)/b)^{1/2}
\end{eqnarray}
where $a=\log(2N/\delta)$, $b=2(N-1)$.  We now derive a variational analogue of Eq. \ref{eq:meta22} based on the techniques used in Theorem 1 above:

 \textbf{Theorem 4} (Variational Meta-Learning Bound). \textit{Using the notation above, with variational distributions represented by DPPs, $r_1:(X,Y)\rightarrow (Z^d)'$, $r_2:\Theta_1\rightarrow (Z^d)'$ and $r_3:\Theta_1\rightarrow (Z^d)'$, the following forms a valid h-o generalization bound, under the condition $(S_1,...,S_M) \indep S$:}
\begin{eqnarray}\label{eq:meta4}
\phi^{\text{ML}}(S_1,...,S_M)&=&\lambda (f_1^\rho,S_{M+1}).\phi^1_b(f_1^{\rho,M+1},S_{M+1};f_2^\pi=\mathcal{A}(S_1,...,S_M)) \nonumber\\
\mathcal{A}(S_1,...,S_M)&=&\text{argmin}_{f_2^\rho}\min_{f_1^{\rho,1...M},r_1,r_2,r_3} \phi^2_b(f_2^\rho, f_1^{\rho,1...M},r_{1...3}) \nonumber\\
\phi^2_b(f_2^\rho, f_1^{\rho,1...M},r_{1...3}) &=& \mathbb{E}_{t}[-\mathbb{E}_{S_t,r_1(\gamma|x,y)}[\log(f_1^{\rho,t}(y|x,\gamma)]+ \mathbb{E}_{S_t}[\text{KL}(r_1(\gamma|x,y),z_1)] + \nonumber\\
&& \mathbb{E}_{f_1^\pi\sim f_2^\pi}((\text{KL}'(f_2^\rho,f_2^\pi)+\mathbb{E}_{z_1(\gamma)f_1^{\rho,t}(\theta_0|\gamma)}[\log z_1(\gamma)+\log f_1^{\rho,t}(\theta_0|\gamma) - \nonumber\\ 
&& \log r_2(\gamma|\theta_0)]  - \mathbb{E}_{f_1^{\rho,t}}[\log(f_1^\pi(\theta_0))] + a)/b)^{1/2}]  +  \nonumber\\ 
&&((\text{KL}'(f_2^\rho, f_2^\pi)+c)/d)^{1/2}  \nonumber\\ 
\text{KL}'(f_2^\rho, f_2^\pi) &=&  \mathbb{E}_{z_2(\gamma)f_2^\rho(\theta_1|\gamma)}[\log z_2(\gamma)+\log f_2^\rho(\theta_1|\gamma) - \nonumber\\ 
&& \log r_3(\gamma|\theta_1)]  - \mathbb{E}_{f_2^\rho}[\log(f_2^\pi(\theta_1))]
\end{eqnarray}
\textit{where we use the same notational shorthands as in Theorem 1, $a=\log(2MN_m/(\delta/(2M)))$, $b=2(N_m-1)$, $c=\log(M/(\delta/2))$ and $d=2(M-1)$.}

{\em Proof.} Similarly to Theorem 1, we first expand out the KL terms in Eq. 5 from the main paper into cross-entropy and negative entropy terms.  We then upper-bound the Gibbs risk with the ELBO bound as in Theorem 1 (introducing $r_1$), and upper-bound the negative entropy terms using the bound $\mathbb{H}(q(x))\geq -\mathbb{E}_{q(x,\gamma)}[\log q(\gamma)+\log q(x|\gamma) - \log r(\gamma|x)]$ from [16].  For this purpose, we introduce variational distribution $r_2$ for the $\mathbb{H}(f_1^{\rho,t})$ terms, and $r_3$ for the $\mathbb{H}(f_2^{\rho})$ terms.  Eq. \ref{eq:meta4} then results from substituting the upper-bounds on these terms into Eq. 5 from the main paper.
\begin{flushright}
$\square$
\end{flushright}

We note that a Th. 4 shares the variational distributions $r_1$ and $r_2$ between tasks; a tighter bound may be derived by introducing distributions $r_{1,1...M}$ and $r_{2,1...M}$ to allow these distributions to vary by task (or by including the task variable as an additional input, hence amortizing the family of distributions).

\section{Second-order Complexity Bounds}\label{app:C}

Below, we provide a proof of Theorem 3 from Sec. 3.2 of the main paper:

\textbf{Theorem 3} (Second-order Complexity ML-Bound). \textit{With notation as in Theorem 2, and assuming $(S_1,...S_M,S'_1,...S'_{M+1})\indep  (f_1,S_{M+1})$, we have the h-o bound:}
\begin{eqnarray}\label{eq:cmplxML}
\phi^{\text{2o-cplx-ML}}(S_{1:M},\mathcal{A}_f)&=&\lambda (f_1,S_{M+1}).\phi^{\text{2o-cplx}}(g=\text{sample}(\mathcal{A}_g(S_{1:M})),\tau,S'_{M+1})(f_1,S_{M+1})\nonumber\\
\mathcal{A}_g(S_{1:M})&=&\text{argmin}_{g_2} \mathbb{E}_{t}[\phi^{\text{2o-cplx}}(g=\text{sample}(g_2),\tau,S'_t)(\mathcal{A}_f(S_{t}),S_{t})] + \eta(g_2)\nonumber\\
\eta(g_2)&=&\frac{1}{\lambda}\left(\text{KL}(g_2,\pi_2)+\log\left(\frac{1}{\delta}\right)+\left(\frac{\lambda^2}{M}\right)\right),
\end{eqnarray}
\textit{where $g_2:G_2=G'_1$, $\mathcal{A}_f:(X\times Y)\rightarrow F_1$, and each task $t$ has its own auxiliary data samples, $S'_{t,1:N'_t}$.  Further, a bound on the transfer error is provided by $\mathbb{E}_{t}[\phi^{\text{2o-cplx}}(g=\text{sample}(g^*_2),\tau,S'_t)(\mathcal{A}_f(S_t),S_{t})] + \eta(g_2)$, where $g^*_2=\mathcal{A}_g(S_1,...,S_M)$.}

{\em Proof.}  We first note that, since the conditions of the theorem ensure $S'_{M+1}\indep(f_1,S_{M+1})$, the bound returned by Eq. \ref{eq:cmplxML} for a new task satisfies the independence conditions of Theorem 2, and hence by the proof of Th. 2 it forms a valid h-o generalization bound.  For the transfer error bound in Th. 3, we note that each term $\phi^{\text{2o-cplx}}(g=\text{sample}(g^*_2),\tau,S'_t)(\mathcal{A}_f(S_t),S_{t})]$ is a valid bound on the task-specific risk for task $t$, since $f_{1,t}$ is chosen by a predefined algorithm $\mathcal{A}_f(S_t)$, which ensures that $S'_{t}\indep(f_{1,t})$.  We can view these bounds themselves as a random variable, with one observation for each of the $M$ tasks.  Hence, applying Eq. \ref{eq:pac_bayes1b} results in the bound on the transfer error noted in the theorem, i.e. the value $\phi^{\text{2o-cplx}}$ takes on a new task, when $g_1$ is sampled according to $g^*_2$, and $f_1$ is set using $\mathcal{A}_f(S_{M+1})$ is with probability $(1-\delta)$ less than $\mathbb{E}_{t}[\phi^{\text{2o-cplx}}(g=\text{sample}(g^*_2),\tau,S'_t)(\mathcal{A}_f(S_t),S_{t})] + \eta(g_2)$.  By the union bound, this holds with probability $1-2\delta$, since it requires both that $\phi^{\text{2o-cplx}}$ returns a value bounded by this quantity on the new task, and that the true risk on the new task does not exceed the returned value.
\begin{flushright}
$\square$
\end{flushright}

The bounds in Theorems 3 and 4 use stochastic generalization classifiers, with the types $g_0:G_0=(F_0\rightarrow\{0,1\})$, $g_1:G_0'$ and $g_2:G_0''$.  For completeness, below we give the explicit forms of these classifiers used in the experimentation:
\begin{eqnarray}\label{eq:stoch_class_g}
g_0 &=& \text{NN}_{F_0,\{0,1\}}(.;\theta_0) \nonumber \\
g_1 &=& \text{NN}_{F_0,\{0,1\}}(.;\text{NN}_{Z^d,\Theta_0}(z_1^+;\theta_1)+e_1^+) \nonumber \\
g_2 &=& \text{NN}_{F_0,\{0,1\}}(.;\text{NN}_{Z^d,\Theta_0}(z_1^{++};\text{NN}_{Z^d,\Theta_1}(z_2^+,\theta_2)+e_2^+)+e_1^{++}),
\end{eqnarray}
where the parameter spaces $\Theta_0$, $\Theta_1$ are as in Eq. 2 from the main paper.  In practice, the input to $g_0$ is a parameter vector $\theta_0:\Theta_0$; we only need consider $f_0$'s which can be represented in the form $\text{NN}_{X,Y}(.;\theta_0)$ in defining $g_0$, since $f_1$, as defined in Eq. 2 from the main paper, returns classifiers only of this kind (the output on other members of $F_0$ can be set arbitrarily).

\end{document}